\pdfoutput=1

\documentclass[11pt]{article}

\usepackage{emnlp2021}

\usepackage{times}
\usepackage{latexsym}
\usepackage{xspace}

\usepackage{soul}
\usepackage{xcolor}
\usepackage{graphicx}
\usepackage{comment}
\usepackage{multirow}
\usepackage{subfigure}

\usepackage[T1]{fontenc}

\usepackage[utf8]{inputenc}

\usepackage{microtype}

\newcommand{\ourtask}{\textsc{KPA-2021}\xspace}
\newcommand{\matchTrack}{\textit{Matching Track}\xspace}
\newcommand{\genTrack}{\textit{Generation Track}\xspace}
\newcommand{\argq}{IBM-ArgQ-Rank-30kArgs\xspace}
\newcommand{\argkp}{ArgKP\xspace}
\newcommand{\argExt}{ArgKP-2021\xspace}
\newcommand{\routineVaccines}{\textit{Routine child vaccinations should be mandatory}\xspace}
\newcommand{\socialMedia}{\textit{Social media platforms should be regulated by the government}\xspace}

\newcommand{\pavincent}{pa\_vincent\xspace}
\newcommand{\mozhiwen}{mozhiwen\xspace}

\newcommand{\kp}{KP\xspace}
\newcommand{\kps}{KPs\xspace}

\newcommand{\mspl}{SMatchToPR\xspace}
\newcommand{\heinrich}{ModrnTalk\xspace}
\newcommand{\vund}{NLP@UIT\xspace}
\newcommand{\viethoang}{MatchTStm\xspace}
\newcommand{\sohanpat}{Enigma\xspace}
\newcommand{\emanuele}{poledro\xspace}
\newcommand{\peratham}{XLNet\xspace}
\newcommand{\teamj}{SiamBERT\xspace}
\newcommand{\moreo}{RoBERTa-LC\xspace}
\newcommand{\barh}{BarH\xspace}

\usepackage[draft]{todonotes}   

\title{Overview of the 2021 Key Point Analysis Shared Task}

  \author {
 Roni Friedman\thanks{\ \ These authors equally contributed to this work.}\hspace{0.15cm},\hspace{0.2cm}Lena Dankin$^{*}$, Yufang Hou, Ranit Aharonov,\\ 
 \textbf{Yoav Katz and Noam Slonim}\\
 IBM Research\\
 \texttt{\{roni.friedman-melamed,lenad,ranit.aharonov2,katz,noams\}@il.ibm.com}\\
\texttt{yhou@ie.ibm.com}}

\begin{document}
\maketitle
\begin{abstract}
We describe the 2021 Key Point Analysis (\ourtask) shared task on key point analysis that we organized as a part of the
8th Workshop on Argument Mining (ArgMining 2021) at EMNLP 2021. We outline various approaches and discuss the results of the
shared task. We expect the task and
the findings reported in this paper to be
relevant for researchers working on text summarization and argument mining.
\end{abstract}

\section{Introduction}

Key Point Analysis (KPA) was introduced in \citet{bar-haim-etal-2020-arguments, bar-haim-etal-2020-quantitative} as a challenging NLP task with tight relations to Computational Argumentation, Opinion Analysis, and Summarization, 
and 
with many practical applications \cite{bar-haim-etal-2021-every}. Given a potentially large collection of relatively short, opinionated texts focused on a topic of interest, the goal of KPA is to produce a succinct list of the most prominent key-points (\kps{}) in the input corpus, along with their relative prevalence. Thus, the output of KPA is a bullet-like summary, with an important quantitative angle. Successful solutions to KPA can be used to gain better insights from public opinions as expressed in social media, surveys, and so forth, giving rise to a new form of a communication channel between decision makers and people that might be impacted by the decision.

Various requirements govern the value of the KPA output. \kps{} are expected to be succinct, non-redundant, capturing points that are central to the topic of interest, and reflecting a clear stance towards that topic. Ideally, they should be at the right granularity for summarising the input data -- not too specific and yet still informative and not overly general. In addition, accurate mapping of input texts to \kps{} is obviously essential. First, to ensure a reliable estimate of the prevalence of each key point. And second, to enable the user to drill-down, to gain a deeper understanding of the issues underlying each key point, as expressed by the input texts mapped to that key point. 

The goal of the \ourtask shared task was to further increase the attention of the NLP community to this emerging task, while enriching the space of existing KPA solutions. Since providing a complete KPA solution is challenging, we divided the task into two tracks, enabling teams to participate only in the first, relatively simpler track. Specifically, in the first track, referred to as \matchTrack, \kps{} are given as part of the input, and the task is focused on mapping input text to these \kps{}. In contrast, in the second track, referred to as \genTrack, no \kps{} are provided and the task requires to further generate the \kps{}. The data being considered for evaluation on both tracks are crowd sourced arguments on three debatable topics. 

The nature of the KPA task requires to consider various evaluation measures to estimate the quality of the results across different dimensions. In this paper we report and analyze the results of $22$ models submitted by $17$ teams to the \ourtask shared task across these different evaluation measures. We also discuss lessons learned from these multiple submissions, that can guide future development of new KPA algorithms and associated evaluation methods. 

\section{Related Work}
\label{sec:related_work}
Our shared task is based on \newcite{bar-haim-etal-2020-arguments, bar-haim-etal-2020-quantitative}, which 
introduced 
the problem of KPA and proposed an extractive approach to address the task. In particular, we
focus on benchmarking two sub tasks: (1) Key point matching; 
and (2) Key point generation and matching. 

For the matching task \newcite{bar-haim-etal-2020-quantitative} uses RoBERTa \cite{Liu2019RoBERTaAR} based classifier, trained on argument-\kp{} pairs. For the generation task \cite{bar-haim-etal-2020-quantitative} employs the following steps:  First, compiling a list of high quality arguments that serve as \kp{} candidates. Next, a match confidence score is calculated between all 
\kp{} candidate-argument pairs, as well as between all the \kp{} candidates to themselves. Finally, several filtering and ordering strategies are employed to maintain the most salient, yet non-overlapping \kp{} candidates to serve as \kps{}. 

Below we briefly review the work closely related to KPA in a few fields.

\paragraph{Multi-document summarization (MDS).} 
Most prior 
work on MDS has been focused on the news, Wikipedia, and scientific literature domains \cite{Dang2005OverviewOD,10.5555/3016100.3016261,j.2018generating,lu-etal-2020-multi-xscience}. 
MDS techniques are typically
query-based and extractive, where 
models select a set of non-redundant sentences that are most relevant to the query to include in the summary \cite{xu-lapata-2020-coarse}. A few studies \cite{j.2018generating,shapira-etal-2021-extending} have explored generating the most salient queries the users might be interested in. However, it is still unclear how these salient queries/points are represented in the source documents. 
KPA 
addresses this problem by reporting 
the prevalence of each 
\kp{} 
in the input data. 

\paragraph{Opinion and argument summarization.} Recently there has been an increasing interest in summarizing opinions expressed in various reviews \cite{amplayo-lapata-2021-informative,elsahar-etal-2021-self} or argumentative text \cite{wang-ling-2016-neural,chen-etal-2019-seeing,syed-etal-2020-news}. KPA contributes to this line of work by adding a quantitative dimension which reflects the distribution of opinions 
in the examined data. 
For a detailed discussion of the relation between KPA and argument clustering and summarization see \cite{bar-haim-etal-2020-arguments,bar-haim-etal-2020-quantitative}.

\section{Task Details}

\subsection{Task Description}

Given a collection of arguments, with either pro or con stance towards a certain debatable topic, the 
goal of the \ourtask task is to generate a KP-based quantitative summary. 
The \genTrack 
requires models to perform the entire task 
-- 
i.e., generate a list of \kps{} for each topic and stance, and then 
predict for each argument in this topic and stance
the confidence by which it is matched to each \kp{}. 
Since this is rather challenging 
we also established the \matchTrack,
in which a set of \kps{}, generated by an expert per topic and stance, is given as part of the input, requiring models to only predict a match confidence score for each argument in the input, towards each of the 
provided \kps{}. 
In both tracks, the confidence score expresses the model's confidence that the \kp{} represents the essence of the argument, and can be associated with it for the purpose of a KP-based quantitative summary. 
Since in the ground truth data (defined in the next section), 
only $5\%$ of the arguments map to multiple \kps{}, we take a simplifying approach 
in the evaluation, where we consider for each argument 
only the \kp{} with the highest confidence score 
as true positive. 

Since in the ground truth data, a significant portion of the arguments are not matched to any \kp{}, 
it is not enough for a model to rank the \kps{} per argument. 
Rather, it should also provide a threshold on the confidence that determines whether the highest scoring \kp{} is matched to the argument, or whether this argument has no matching \kp{}. In this shared task, we opted for a simpler approach, in which the submitted models are not required to determine the threshold. Rather, we evaluate all models by considering only the top $50\%$ of the argument-\kp{} pairs, as ranked by 
the confidence scores of all pairs. 

The full terms of the \ourtask task can be found online\footnote{\url{https://github.com/ibm/KPA\_2021_shared_task}}.

\subsection{The \argExt data set}

For \ourtask we use a data set covering  
a set of debatable topics, where for each topic and stance, 
a set of triplets of the form <argument, \kp{}, label> is provided. 
The data set is based on the \argkp data set \citep{bar-haim-etal-2020-arguments}, which 
contains arguments contributed by the crowd on $28$ debatable topics, split by their stance towards the topic, and 
\kps{} written by an expert 
for those topics.
As described in \citep{bar-haim-etal-2020-arguments}, crowd annotations were collected to determine whether a \kp{} represents an argument, i.e., is a match for an argument, and based on these annotations every argument-\kp{} pair received a label that can take three values: (i) positive - $60\%$ or more of the annotations determined there is a match; (ii) negative - $15\%$ or less of the annotations determined there is a match; 
and (iii) undecided - neither positive nor negative. For example, in the pro stance towards the topic "Homeschooling should be banned", and for the \kp{} "Mainstream schools are essential to develop social skills", the argument "children can not learn to interact with their peers when taught at home" was labeled as positive and the argument "homeschooling causes a decline in education" as negative.
The arguments in \argkp are a subset of the \argq data set
\citep{Gretz_Friedman_Cohen-Karlik_Toledo_Lahav_Aharonov_Slonim_2020}, which consists of $~30$K crowd-sourced arguments on $71$ controversial topics, where each argument 
is labeled for its stance towards the topic as well as for its quality. 

The \argkp data set, which is publicly available, served as the training data for our task. The $28$ topics in the data set were split to train ($24$ topics) and dev ($4$ topics) sets and were offered to the participants with a recommendation to use as such in the preparation of their models. 
For \genTrack, in addition to the \argkp data set, the participants were encouraged to use the remaining arguments from the \argq data set.

For a test set, we extended \argkp, adding three new debatable topics, 
that were also not part of \argq. 
The test set was collected specifically for \ourtask, and was carefully designed to be similar in various aspects to the training data\footnote{Those aspects include, among others -- number of arguments in each topic; the distribution of their quality and length; the amount and style of expert \kps{} produced for them; as well the coverage of those as obtained by crowd labeling.}. For each topic, crowd sourced arguments were collected, expert \kps{} generated, and match/no match annotations for argument/\kp{} pairs obtained, resulting in a data set compatible with the \argkp format.   
Arguments collection strictly adhered to the guidelines, quality measures, and post processing used for the collection of arguments in \argq \cite{Gretz_Friedman_Cohen-Karlik_Toledo_Lahav_Aharonov_Slonim_2020}, while the generation of expert \kps{}, collection of match annotations, and final data set creation strictly adhered to the manner in which \argkp was created \cite{bar-haim-etal-2020-arguments}.  

The full data set, i.e., \argkp together with the test set, is termed \argExt and is available online\footnote{\url{https://www.research.ibm.com/haifa/dept/vst/debating\_data.shtml}}. 
See Table \ref{data:stats} for data statistics. 

\begin{table*}[ht!]
\begin{center}
\begin{tabular}{c|c|c|c|c|c |c|c} 
 \hline
 Split & \#Args & \#KPs  & Positive  & Negative & Undecided & Strict  coverage & Relaxed coverage \\ \hline
train (24) & 5583 & 207 & 0.17 &  0.67 & 0.16 &0.72 & 0.94\\
dev (4) & 932 & 36 & 0.18 &  0.65 & 0.18 &0.72 & 0.95\\
test (3) & 723 & 33 & 0.14 &  0.73 & 0.13 &0.69 & 0.91\\

\end{tabular}
\end{center}
\caption{The \argExt data set. The number of topics in each split is given in brackets. 
Coverage represents the  
fraction of arguments that have at least one matching \kp{}. 
The strict view considers only argument-\kp{} pairs with a positive label, while the relaxed view considers all pairs with a positive or undecided label. See text for details.
\label{data:stats}
}
\end{table*}

\subsection{Evaluation Metrics}
\label{sec:evaluation}
\subsubsection{\matchTrack}

For the submitted prediction files, we extract argument-\kp{} pairs as follows. 
First, each argument is paired with the highest scoring \kp{} assigned to it (randomly chosen in case of a tie)\footnote{In such cases, each experiment was repeated 10 times with different random seeds, and we report the resulting average. Standard deviation for those rare cases is insignificant.}. 
Second, the top $50\%$ of the argument-\kp{} pairs are extracted for each topic and stance as an ordered list, based on the confidence score.
Finally, given the ground truth data, Average Precision (AP, 
\citet{mAP_reference}), is calculated per topic and stance. Note that a perfect score of $1$ can only be achieved when all top $50\%$ matches are correct.
We chose AP as the evaluation metric, since it supports evaluating the correlation between a model's confidence and prediction success. 
We obtain the mean AP (mAP) by 
macro 
averaging over all topics and stances 
so that each topic and stance has the same effect over final score. 

When evaluating the models, we need to  
address the issue of the undecided argument-\kp{} pairs in the labeled data. 
Considering a prediction of undecided pairs as a false prediction is too strict, since $>15\%$ of the annotators found this match correct, and hence it is not entirely wrong. 
Discarding 
predictions of undecided pairs from the set of a model's predictions, as was done in \cite{bar-haim-etal-2020-arguments}, is not relevant for \ourtask, as it would have rendered the evaluation process inconsistent 
across different models. 
We therefore relied on 
two measures for this track - \textbf{relaxed} and \textbf{strict} mAP. In the strict evaluation score, undecided pairs are considered as wrong predictions, while in the relaxed evaluation score they are considered as correct, 
i.e., true positives. 
The final rank of a model in this track is the average between the rank of the model in the strict score and its rank in the relaxed score.

\subsubsection{\genTrack}
Models submitted to this track were evaluated on two aspects: The first is how well 
the set of generated \kps{} serves 
as a summary of possible arguments for this topic and stance. 
The second is how well does the model match between the arguments and the generated \kps{}.
We note that evaluating only how well the arguments match the generated \kps{} is insufficient since 
one may obtain relatively high mAP scores by generating redundant and/or very general and non-informative \kps{}. It is therefore crucial to establish the quality of the set of \kps{} as a standalone 
evaluation.
Due to the nature of this track, no ground truth data is available, and we therefore 
perform the evaluation of the submitted results using crowd annotations via the Appen platform\footnote{www.appen.com.}. 
The annotations were generated by a selected group of annotators which performed well in previous tasks, 
and further obtained good results in 
sanity-test 
questions and demonstrated 
high agreement with other annotators (\citet{Gretz_Friedman_Cohen-Karlik_Toledo_Lahav_Aharonov_Slonim_2020}, \citet{toledo-etal-2019-automatic}).

The first evaluation considers the model's ability to correctly match arguments to \kps{}. As in \matchTrack{}, an argument is matched to a single \kp{}, and only the top $50\%$ matches are considered for evaluation. 
Unlike \matchTrack, here the evaluation of the match must be retrospective, since the \kps{} are provided by the submitted models.
Following \citet{bar-haim-etal-2020-quantitative}, we presented each argument to the annotators with its top matching \kp{}, and asked if the \kp{} matches the argument. Possible answers are 'Yes'/'No'/'Faulty \kp{}'. We consider 'Faulty \kp{}' as no match. Since here we directly ask regarding the predicted match, rather than showing annotators all possible pairs (as was done when generating the ground truth data for the \matchTrack{}), we generate a binary label from the annotations by the majority vote, and compute a single mAP 
score. 

The second evaluation aspect concerns the quality of the set of \kps{} generated by the submitted model. Here we asked annotators to perform a pairwise comparison between two sets of \kps{}, each taken from a different model, on the same topic and stance. We perform such pairwise comparisons between all pairs of top  models according to \genTrack mAP score.
Each \kp{} set is sorted, such that the more prevalent 
\kps{} according to this model's matching are 
higher in the list. 
\kps{} that match $1\%$ or less of the arguments are not presented to the annotators. 

The annotators are first asked to determine 
for each \kp{} set what is the stance of the set towards the topic (All support the topic/All contest the topic/Most support the topic/Most contest the topic). This initial question is not positioned as a comparative question. However, annotators' answers are compared to the stance of the arguments of the \kp set, and in case one model is correct and the other is not (either all or some \kps{} are judged not to be in the correct stance), this is counted as a win for the former. Two comparative questions follow: \textit{Which of the two summaries is less redundant?} and \textit{Which of the two summaries captures points that are more central to the topic?}, where the latter question aims for the relevance of the \kp{} set. 
Possible answers for those questions are \textit{Summary A/Summary B/Both are equal in this respect}. 
For each of the three questions, ranking is calculated using the Bradley-Terry model \cite{bradley_terry}, which 
predicts the probability of a given participant to win a paired comparison, based on previous paired comparison results of multiple participants, and thus allows ranking them.
The final rank of a model in the \kp{} set quality aspect is the average of the three ranks.

Finally, a model's rank in the \genTrack{} is the average of its rank in the mAP evaluation aspect and its rank in the evaluation of the \kp{} set quality. In case of a tie between the two ranks, mAP score is given preference. 

\section{Submitted models}

In total, 17 models were submitted to \matchTrack, and 5 models to \genTrack. 
Note that some teams did not provide description for their model. 
Still, we
show their results in the results section, but we cannot provide their descriptions.

\subsection{\matchTrack} 
As in other NLP related tasks, transformer based approaches are preferable among the participants, with a clear dominance of RoBERTa \cite{liu-lapata-2019-roberta}. None of the participants found the undecided pairs as 
useful for training, thus 
training was focused only on 
positive and negative pairs. Most of the participants used the topic as an input to their models, 
and 
ignored the stance. 
We also note that some of the models (\mspl, \viethoang) utilized contrastive learning, taking advantage of the fact that arguments are provided with matched and unmatched \kps{}.  \textbf{\teamj}, though not directly by contrastive learning, also shows two key points for one argument in training. 

\textbf{SMatchToPageRank} (hereafter as \textbf{\mspl}) \cite{KPA2021-18} aims to learn an embedding space where matching argument-\kp{} pairs are closer than non-matching pairs. This model has two inputs: (1) the argument; and (2) the concatenation of the \kp{} and the topic. The tokens of each input are embedded using RoBERTa-large. Next, those inputs are passed to a siamese Sentence-BERT network that is trained with a contrastive loss. 

Team \textbf{\vund} use ALBERT XXLarge \cite{DBLP:journals/corr/abs-1909-11942} pre-trained model for sentence-pair classification for pairs of argument and \kp{}. Five models are fine-tuned on different cross-validation splits, and for inference the average of all models is considered as the final score.

\textbf{Modern Talking - RoBERTa-Base} (hereafter as \textbf{\heinrich}) \cite{KPA2021-6} is a RoBERTa-Base model fine-tuned on the concatenation of arguments-kps{} pairs. At training, the model is warmed up on 6\% of the data.

\textbf{\sohanpat} \cite{KPA2021-31} is based on a DeBERTa-Large \cite{he2020deberta} pre-trained model. The input to the model is the concatenation of \kp{}, argument, and topic. The output of the DeBerta model is concatenated with the POS tags representation of the three text pieces, and then fed to two more dense layers. 

\textbf{Matching the statement} (hereafter as \textbf{\viethoang}  \cite{KPA2021-17} embeds each argument and \kp{} separately along with their context that is represented by the topic and stance. RoBERTa is utilized to extract all text embeddings. Next, the statement pairs are passed to a fully connected network. Finally, a contrastive learning is employed by grouping matching and non matching arguments for the same \kp{} in a single batch. 

\textbf{\emanuele} \cite{KPA2021-27} is a fine-tuned 
ALBERT-XXLarge 
model, similarly to \vund. The input to the model is a concatenated argument-\kp{} pair. 

\textbf{\peratham} is a vanilla XLNet-Large auto-regressive model \cite{NEURIPS2019_XLNet} fine-tuned 
on argument-\kp{} pairs. 

\textbf{RoBERTa-large-concat} (hereafter as \textbf{RoBERTaLC}) is a fined tuned RoBERTa-large model. The input to the model is the concatenation of the topic, the \kp{}, and the argument. Training data is balanced between positive and negative match examples by over sampling positive instances. 

\textbf{Siamese BERT classifier} (hereafter as \textbf{\teamj}) is a siamese BERT \cite{devlin-etal-2019-bert} based model trained on pairs of arguments and \kps{}. 
At each training step, the same argument with two different \kps{}, 
one that matches and one that doesn't, is passed to the model, and it learns to prefer the matching pair.

\subsection{\genTrack} 

\textbf{\mspl} applies an extractive approach. First, arguments with high quality scores (using \cite{toledo-etal-2019-automatic, Gretz_Friedman_Cohen-Karlik_Toledo_Lahav_Aharonov_Slonim_2020}) 
are selected as \kp{} 
candidates. Next, arguments with high matching score (using \matchTrack models) are connected to form an undirected graph, where the nodes are the arguments. Finally, an importance score is calculated for each node, and most important nodes are selected as \kps{}. Note that \kp{} candidates that have high confidence scores with other candidate are ignored, to ensure diversity. 

\textbf{\sohanpat} approaches this task as an abstractive summarization task. First, Pegasus \cite{pmlr-v119-zhang20ae} is fine-tuned with an input of concatenated arguments and topics. The set of corresponding \kps{} 
are provided as a ground truth summary. For each pair of arguments and the corresponding topic, several summaries were generated. Next, those summaries are ranked by the  ROUGE-1 score with 
respect to the input pair, and the top 5 summaries are selected as \kps{} for each stance.

\textbf{\peratham} applies the XLNet model developed for \matchTrack on all possible pairs of arguments, so that each argument is evaluated as a 
potential 
\kp{} summarizing another argument. Arguments with the highest average confidence scores are selected as \kps{}. 
\section{Results}
In addition to the submitted models, we evaluate the 
state of the art RoBERTa based model described in \cite{bar-haim-etal-2020-quantitative} and in section \ref{sec:related_work}, termed hereafter \barh{}. 

\subsection{\matchTrack}

\begin{figure*}[ht!]
\centering
  \subfigure[]{\includegraphics[scale=0.30]{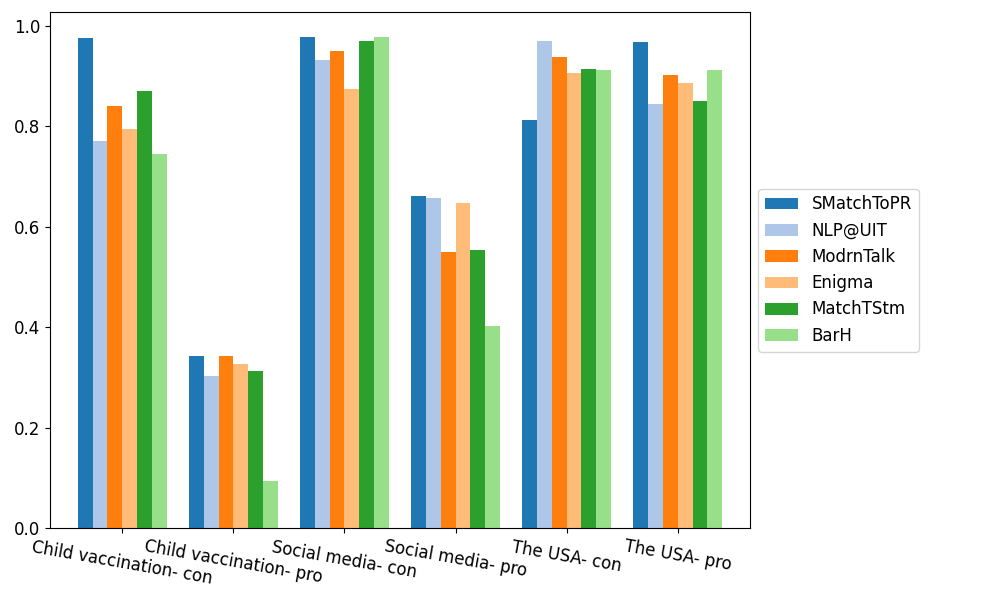}}
  \subfigure[]{\includegraphics[scale=0.30]{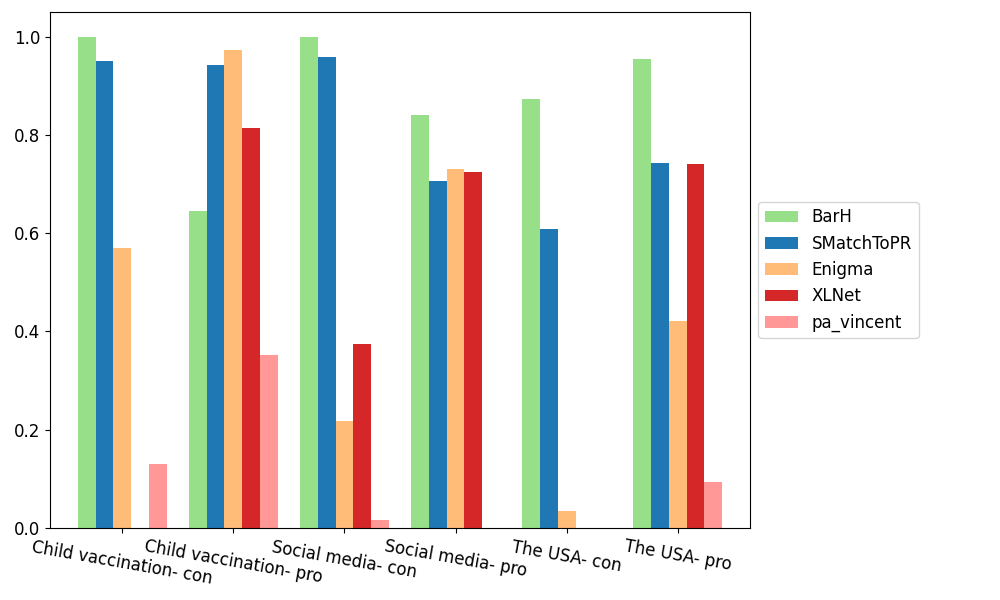}}
\caption{AP per topic and stance for the top models and \barh. (a) \matchTrack. (b) \genTrack. Missing bars mean a value of 0.
\label{fig:mAP_per_motion}}
\end{figure*}

Table \ref{results:track1_results_full} describes the evaluation results for the \matchTrack. By definition, the relaxed mAP is higher than the strict 
mAP. Still, it is interesting to note that many models achieve a high relaxed mAP, while differences between models are more evident when considering the strict mAP.
The top-ranked model in this track, \mspl{}, is 
also ranked first when considering the average \textit{value} of the mAP scores, as well as p@50\%. 
However, the other models tend to swap positions when considering different measures, though there is a distinct group of high performing models. 
Figure \ref{fig:mAP_per_motion}(a) depicts strict mAP scores for each topic and stance, for the top performing models and for \barh. 
Evidently, some topic/stance pairs are easier than others for all models. 

\begin{table*}[ht!]
\begin{center}
\small
\begin{tabular}{c| c| c| c || c | c} 
 \hline
Model & Rank &Strict mAP (r) & Relaxed mAP (r) & Mean of mAP (r) & Strict p@50\% (r)\\ \hline
\mspl * & 1 & 0.789 (1) & 0.927 (4) & 0.858 (1) & 0.848 (1) \\
\vund * & 2 & 0.746 (3) & 0.93 (3) & 0.838 (2) & 0.827 (3) \\
\heinrich * & 3 & 0.754 (2) & 0.902 (6) & 0.828 (4) & 0.806 (5) \\
\sohanpat * & 4 & 0.739 (5) & 0.928 (4) & 0.833 (3) & 0.828 (2) \\
\viethoang * & 5 & 0.745 (4) & 0.902 (6) & 0.824 (5) & 0.808 (4) \\
mozhiwen & 5 & 0.681 (9) & 0.948 (1) & 0.814 (6) & 0.784 (7) \\
fengdoudou & 7 & 0.676 (10) & 0.944 (2) & 0.81 (7) & 0.753 (13) \\
\emanuele * & 8 & 0.717 (6) & 0.901 (8) & 0.809 (8) & 0.799 (6) \\
Fibelkorn & 9 & 0.706 (7) & 0.872 (12) & 0.789 (9) & 0.779 (9) \\
\peratham * & 10 & 0.684 (8) & 0.869 (13) & 0.777 (10) & 0.765 (11) \\
\moreo * & 11 & 0.661 (13) & 0.885 (10) & 0.773 (11) & 0.774 (10) \\
pa\_vincent & 12 & 0.596 (15) & 0.893 (9) & 0.744 (14) & 0.689 (15) \\
\barh * & 13 & 0.674 (11) & 0.865 (14) & 0.77 (12) & 0.78 (8) \\
np-quadrat & 13 & 0.601 (14) & 0.877 (11) & 0.739 (15) & 0.729 (14) \\
niksss & 15 & 0.672 (12) & 0.857 (15) & 0.764 (13) & 0.759 (12) \\
\teamj * & 16 & 0.555 (16) & 0.729 (16) & 0.642 (16) & 0.678 (16) \\
yamenajjour & 17 & 0.539 (17) & 0.682 (17) & 0.61 (17) & 0.667 (17) \\
danielwaechtleruni & 18 & 0.39 (18) & 0.601 (18) & 0.495 (18) & 0.511 (18) \\
\end{tabular}
\end{center}
\caption{\matchTrack{} results. Ranks on each measure are in brackets. 
In addition to the official evaluation metrics, two more scores are provided: the average value of the strict and relaxed mAP values, and p@50\% for the strict view. 
Submissions that provided a descriptions are marked with (*). 
\label{results:track1_results_full}
}
\end{table*}

\subsection{\genTrack}
 
 \begin{table}[ht!]
\begin{center}
\small
\begin{tabular}{c| c| c| c} 
 \hline

Model	 & Rank &  mAP (r) & \kp{} set rank \\ \hline
\barh \* & \centering	1 &	0.885 (1) &	1 / (1,1,2) \\
\mspl \* &	\centering2 &	0.818 (2) &	2 / (2,1,2) \\
\sohanpat \* &	\centering3 &	0.491 (3) &	4 / (2,4,3) \\
\peratham \*	& \centering4 &	0.443 (4) & 3 / (4,3,1) \\
\pavincent	& \centering5 &	0.096 (5) &	None \\
mozhiwen	& \centering6 &	0 (6) &		None \\
\end{tabular}
\end{center}
\caption{\genTrack{} results. \kp{} set rank shows the overall rank and in brackets the rank on each of the three questions: stance, redundancy, relevance.
\label{results:track2_results_full}}
\end{table}

The results of the \genTrack are presented in Table 
\ref{results:track2_results_full} for both the matching aspect (mAP) 
as well as the quality of the generated \kps{}. The full results of the comparative annotation of the sets can be found in Table  \ref{app:bt_table} in the appendix. In general, \barh clearly outperformed all submitted models, while \mspl obtained the best results amongst the submitted models. 
Figure \ref{fig:mAP_per_motion}(b) shows the mAP scores for each topic and stance, for the top performing models and for \barh. 
We note that the inconsistent ranking of \peratham{}, namely first rank on relevance, 
while scoring much worse on other quality questions, stems from the fact that the model provided the set of \kps{} of the positive stance arguments for both the positive and negative stance. Although such a result can achieve high relevance, it fails when considering stance and redundancy, as 
indicated by the results. 

It is interesting to note that \barh outperformed all submitted models by a significant margin, even though it had a relatively low rank in \matchTrack. This could be due to its specific failure in matching to expert \kps{} in one of the topic/stance pairs (see Figure \ref{fig:mAP_per_motion}(a), second set of bars) 
coupled with the usage of macro-averaging. 

When designing the task, we did not direct the participants regarding whether to use an extractive or abstractive approach for generating the 
\kps{}. Indeed, three models, \barh, \mspl, and \peratham used an extractive approach, selecting specific arguments as \kps{}, while three models, \sohanpat, \pavincent, and \mozhiwen opted for an abstractive approach, generating \kps{} de novo. 
Evidently, the extractive approach prevailed. Out of the three models taking the abstractive approach, only  
\sohanpat generated comprehensible \kps{}. The other two, received near zero scores in the matching evaluation, and hence, according the to \ourtask terms, were not included in the \kp{} set evaluation.  

\section{Additional Analysis}

\begin{figure*}[ht!]
\centering

    \subfigure[]{\includegraphics[width=3.1in]{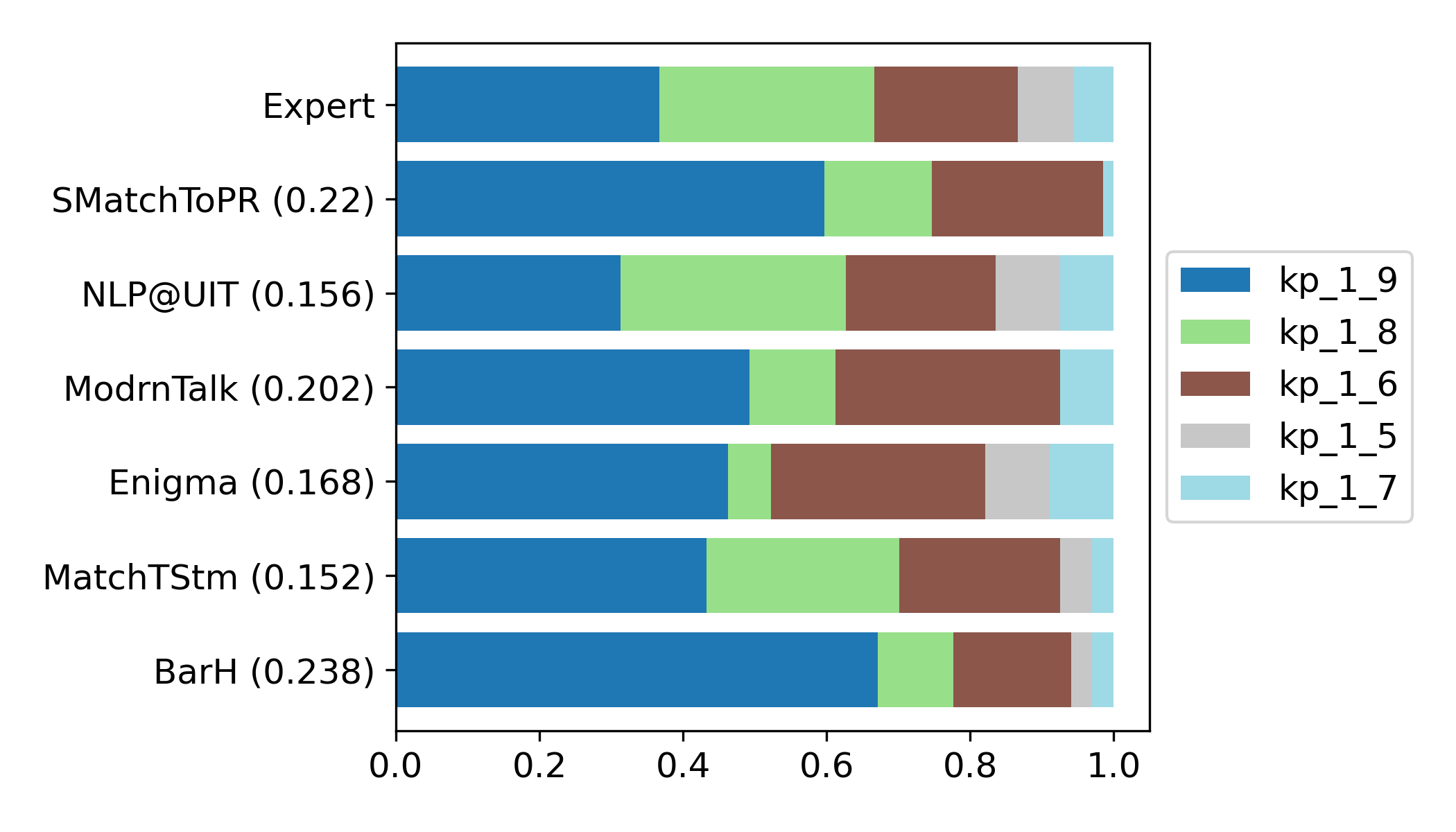}} \hspace{5mm}
    \subfigure[]{\includegraphics[width=2.35in]{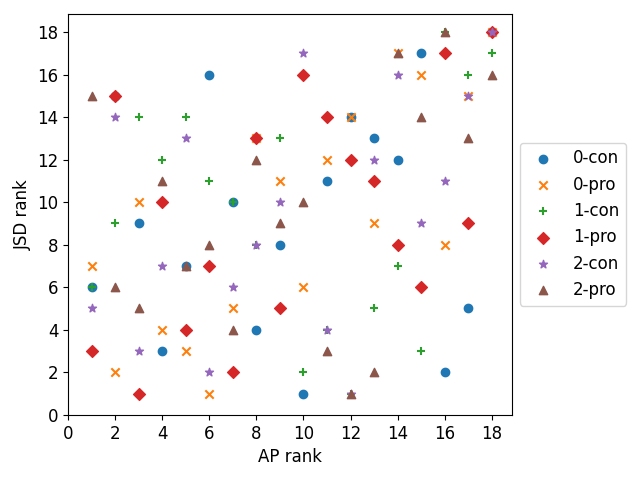}}

\caption{
(a): The distribution of \kps{} among the arguments, in the pro stance on the topic \socialMedia, for the expert ground truth (strict view) and selected models, normalized to the number of arguments that match to a \kp{}. The JSd vs. the expert distribution is given in brackets. (b) JSd rank vs. AP rank for all models in the \matchTrack{}, per topic and stance. The map between topic id, and topic text can be found at Table \ref{tab:test_topics}. Rank 1 means best performance.
\label{fig:distribution_barplot}}
\end{figure*}

Accurate matching of arguments to \kps{} aims to properly reflect the prevalence of each \kp{} in the data to facilitate more informed decision making. 
In this section we analyze how well different models were able to reproduce the presumably true distribution of \kps{} in the input data. To that end, we focus on the \matchTrack{}, and consider the distribution of the expert \kps{} as the ground truth distribution. 
As can be seen in the top row of Figure 
\ref{fig:distribution_barplot}(a), the five \kps{} provided by the expert for this topic and stance are not equally represented in the argument set. 
Nearly 40\% of the arguments were matched by the annotators to "Social media regulation is required to deal with malicious users" (kp-1-9); around 30\% were matched to "Social media regulation can help to deal with negative content" (kp-1-8); and so on. 
Correspondingly, for each model submitted in this track, we compare the distribution of \kps{} as predicted by its matching to the manually obtained ground truth distribution. 
To that end, we use the Jensen Shannon divergence (JSd). 
The smaller the divergence, 
the more accurate 
is the summary generated by a model (based on the 50\% of the arguments that pass the 
threshold). 
We find that the average topic-stance JSd for the top $5$ ranking models ranges from $0.152$ to $0.238$ (see Table \ref{tab:JS_divergence} in 
the 
appendix for more details). 
Interestingly, the models which perform best in the \matchTrack{} evaluation, i.e., have the highest mAP scores, 
are not those with the lowest JSd scores. 
This can also be seen in Figure 
\ref{fig:distribution_barplot}(b). The rank of all models based on their
JSd score 
is plotted vs. their rank in the matching task, for all topics and stances. Although it is evident that models which perform poorly on the matching task 
are also ranked low by JSd, 
the overall correlation is low.

In other words, a model with sub-optimal matching capabilities may still perform relatively well in capturing the overall distribution of the \kps{} in the input data. 

\begin{table}[ht!]
\begin{center}
\small

\begin{tabular}{c|p{6cm}} 
 \hline

Topic id & Text\\
\hline
0 & Routine child vaccinations should be mandatory\\ \hline
1 & Social media platforms should be regulated by the government\\ \hline
2 & The USA is a good country to live in \\ 
\hline
\end{tabular}
\end{center}
\caption{Topics of the test set.
\label{tab:test_topics}}
\end{table}

In the \genTrack{}, there is no single set of \kps{}, since each model generates its own set. It is therefore not relevant to compare the distributions 
over \kps{} as done above. 
However, it is possible to perform a qualitative anecdotal analysis. Table \ref{tab:key_points_full} in the appendix lists the \kps{} generated for one topic and stance, by the expert and the different models.  

Importantly, the expert produced \kps{} 
{\it a priori\/}, 
without seeing the actual set of arguments; while the models generated \kps{} based on the input arguments. Inspection of the generated \kps{} 
suggests that correspondingly, 
some \kps{} were perhaps overlooked by the expert. For example, \barh generated a \kp{} concerning keeping schools safe, and another one concerning prevention of pandemic outbreaks, both of which are not in the expert set, even though there are expert \kps{} that are related.
In Table \ref{tab:small_key_points_list} we show an example of one expert \kp{}, to which we manually matched generated \kps{} from the top $3$ models, as an anecdotal distribution comparison. In this example, \barh and \mspl extracted the exact same \kp{}, yet in \barh it maps to $35\%$ of the arguments, which is in line with the weight of this \kp{} in the expert ground truth, while \mspl maps it to a much larger fraction of the arguments.
 
\begin{table*}[!ht]
\centering
\small
\begin{tabular}{c|p{11.5cm}|c}
\hline
Model & \centering \kp{} & Fraction\\
\hline 
expert & Children should not suffer from preventable diseases & 29.0\\ \hline \hline
\barh &  Vaccines help children grow up healthy and avoid dangerous diseases & 35.0 \\ \hline
 
\mspl &  Vaccines help children grow up healthy and avoid dangerous diseases & 77.0 \\ \hline
\multirow{2}{*}{\sohanpat} & Child vaccinations are necessary to protect against diseases in the child & 50.0\\ \cline{2-3}
& A mandatory vaccination is important to protect against diseases in the child & 25.0  \\ \hline

\end{tabular}
\caption{\kps{} for \routineVaccines in the pro stance, related to the expert \kp{} on the top row. Fraction is out of all arguments with a matched \kp{} (for expert we use the strict view). 
\label{tab:small_key_points_list}}
\end{table*}
\section{Discussion}

We presented \ourtask, the first shared task focused on key point analysis. The shared task received $22$ submissions from $17$ teams, covering both the \matchTrack and the \genTrack. We presented the submitted models, their performance on different evaluation measures, and further analyzed the results.

As expected by its simpler nature, \matchTrack received more submissions. However, evidently, success in this intermediate matching task, does not guarantee success in the overall KPA task, which also requires generating the \kps{}. Future work should determine whether future KPA shared tasks should focus on the \genTrack, or perhaps modify the evaluation measures of the intermediate \matchTrack, such that they better reflect the model performance in the full KPA task. 

Given the inherent subjective nature of matching arguments to \kps{}, we opted for a ternary label, allowing argument-\kp{} pairs to receive an "undecided" label, if the annotators votes were inconclusive. These undecided pairs are eventually considered either as positive or as negative matches, resulting with two gold standards, one potentially too strict, while the other perhaps too lenient. In future tasks, it may be valuable to consider a binary labeling scheme that will give rise to a single gold standard. Such labeling may be created conditionally, such that an undecided pair is marked as a positive match if and only if a minimum number of annotators marked it as positive and no other \kp{} was labeled as matching to that argument. 

A further point for future consideration, is the choice of a pre-defined threshold of 50\%, that guided our evaluation. Although, this has the advantage of not requiring submissions to tune a confidence threshold, it has the limitation that it complicates the evaluation, since the ground truth coverage depends on the arguments, topic and stance. 
A possible alternative would be to require each model to set its own minimum confidence threshold to determine if an input argument should not be matched to any \kp{}. 

Finally, in this task, we used comparative evaluations to determine the quality of the sets of generated \kps{}. Obviously, this results with ranking-based evaluation. Establishing an absolute evaluation metric in this context merits further investigation. 

We thank all participating teams for taking the time to participate in this challenging new shared task. We believe that these contributions, along with 
the data collected and shared in this report, will be valuable to further advance the research around KPA and related tasks.

\section*{Acknowledgments}
We would like to thank Roy Bar-Haim for his helpful comments and suggestions.

\bibliography{anthology,custom}

\begin{thebibliography}{30}
\expandafter\ifx\csname natexlab\endcsname\relax\def\natexlab#1{#1}\fi

\bibitem[{Alshomary et~al.(2021)Alshomary, Gurcke, Syed, Heinisch,
  Spliethöver, Cimiano, Potthast, and Wachsmuth}]{KPA2021-18}
Milad Alshomary, Timon Gurcke, Shahbaz Syed, Philipp Heinisch, Maximilian
  Spliethöver, Philipp Cimiano, Martin Potthast, and Henning Wachsmuth. 2021.
\newblock Key point analysis via contrastive learning and extractive argument
  summarization.
\newblock In \emph{Proceedings of the 8th Workshop on Argument Mining}, Online.
  Association for Computational Linguistics.

\bibitem[{Amplayo and Lapata(2021)}]{amplayo-lapata-2021-informative}
Reinald~Kim Amplayo and Mirella Lapata. 2021.
\newblock \href {https://aclanthology.org/2021.eacl-main.229} {Informative and
  controllable opinion summarization}.
\newblock In \emph{Proceedings of the 16th Conference of the European Chapter
  of the Association for Computational Linguistics: Main Volume}, pages
  2662--2672, Online. Association for Computational Linguistics.

\bibitem[{Bar-Haim et~al.(2020{\natexlab{a}})Bar-Haim, Eden, Friedman, Kantor,
  Lahav, and Slonim}]{bar-haim-etal-2020-arguments}
Roy Bar-Haim, Lilach Eden, Roni Friedman, Yoav Kantor, Dan Lahav, and Noam
  Slonim. 2020{\natexlab{a}}.
\newblock \href {https://doi.org/10.18653/v1/2020.acl-main.371} {From arguments
  to key points: {T}owards automatic argument summarization}.
\newblock In \emph{Proceedings of the 58th Annual Meeting of the Association
  for Computational Linguistics}, pages 4029--4039, Online. Association for
  Computational Linguistics.

\bibitem[{Bar-Haim et~al.(2021)Bar-Haim, Eden, Kantor, Friedman, and
  Slonim}]{bar-haim-etal-2021-every}
Roy Bar-Haim, Lilach Eden, Yoav Kantor, Roni Friedman, and Noam Slonim. 2021.
\newblock \href {https://doi.org/10.18653/v1/2021.acl-long.262} {Every bite is
  an experience: {K}ey {P}oint {A}nalysis of business reviews}.
\newblock In \emph{Proceedings of the 59th Annual Meeting of the Association
  for Computational Linguistics and the 11th International Joint Conference on
  Natural Language Processing (Volume 1: Long Papers)}, pages 3376--3386,
  Online. Association for Computational Linguistics.

\bibitem[{Bar-Haim et~al.(2020{\natexlab{b}})Bar-Haim, Kantor, Eden, Friedman,
  Lahav, and Slonim}]{bar-haim-etal-2020-quantitative}
Roy Bar-Haim, Yoav Kantor, Lilach Eden, Roni Friedman, Dan Lahav, and Noam
  Slonim. 2020{\natexlab{b}}.
\newblock \href {https://doi.org/10.18653/v1/2020.emnlp-main.3} {Quantitative
  argument summarization and beyond: Cross-domain key point analysis}.
\newblock In \emph{Proceedings of the 2020 Conference on Empirical Methods in
  Natural Language Processing (EMNLP)}, pages 39--49, Online. Association for
  Computational Linguistics.

\bibitem[{Baumel et~al.(2016)Baumel, Cohen, and
  Elhadad}]{10.5555/3016100.3016261}
Tal Baumel, Raphael Cohen, and Michael Elhadad. 2016.
\newblock Topic concentration in query focused summarization datasets.
\newblock In \emph{Proceedings of the Thirtieth AAAI Conference on Artificial
  Intelligence}, AAAI'16, page 2573–2579. AAAI Press.

\bibitem[{Bradley and Terry(1952)}]{bradley_terry}
Ralph~Allan Bradley and Milton~E. Terry. 1952.
\newblock \href {https://doi.org/10.1093/biomet/39.3-4.324} {{RANK ANALYSIS OF
  INCOMPLETE BLOCK DESIGNS: THE METHOD OF PAIRED COMPARISONS}}.
\newblock \emph{Biometrika}, 39(3-4):324--345.

\bibitem[{Chen et~al.(2019)Chen, Khashabi, Yin, Callison-Burch, and
  Roth}]{chen-etal-2019-seeing}
Sihao Chen, Daniel Khashabi, Wenpeng Yin, Chris Callison-Burch, and Dan Roth.
  2019.
\newblock \href {https://doi.org/10.18653/v1/N19-1053} {Seeing things from a
  different angle:discovering diverse perspectives about claims}.
\newblock In \emph{Proceedings of the 2019 Conference of the North {A}merican
  Chapter of the Association for Computational Linguistics: Human Language
  Technologies, Volume 1 (Long and Short Papers)}, pages 542--557, Minneapolis,
  Minnesota. Association for Computational Linguistics.

\bibitem[{Cosenza(2021)}]{KPA2021-27}
Emanuele Cosenza. 2021.
\newblock Key point matching with transformers.
\newblock In \emph{Proceedings of the 8th Workshop on Argument Mining}, Online.
  Association for Computational Linguistics.

\bibitem[{Dang(2005)}]{Dang2005OverviewOD}
Hoa~Trang Dang. 2005.
\newblock Overview of duc 2005.
\newblock In \emph{In Proceedings of the Document Understanding Conf. Wksp.
  2005 (DUC 2005) at the Human Language Technology Conf./Conf. on Empirical
  Methods in Natural Language Processing (HLT/EMNLP}.

\bibitem[{Devlin et~al.(2019)Devlin, Chang, Lee, and
  Toutanova}]{devlin-etal-2019-bert}
Jacob Devlin, Ming-Wei Chang, Kenton Lee, and Kristina Toutanova. 2019.
\newblock \href {https://doi.org/10.18653/v1/N19-1423} {{BERT}: Pre-training of
  deep bidirectional transformers for language understanding}.
\newblock In \emph{Proceedings of the 2019 Conference of the North {A}merican
  Chapter of the Association for Computational Linguistics: Human Language
  Technologies, Volume 1 (Long and Short Papers)}, pages 4171--4186,
  Minneapolis, Minnesota. Association for Computational Linguistics.

\bibitem[{Elsahar et~al.(2021)Elsahar, Coavoux, Rozen, and
  Gall{\'e}}]{elsahar-etal-2021-self}
Hady Elsahar, Maximin Coavoux, Jos Rozen, and Matthias Gall{\'e}. 2021.
\newblock \href {https://aclanthology.org/2021.eacl-main.141} {Self-supervised
  and controlled multi-document opinion summarization}.
\newblock In \emph{Proceedings of the 16th Conference of the European Chapter
  of the Association for Computational Linguistics: Main Volume}, pages
  1646--1662, Online. Association for Computational Linguistics.

\bibitem[{Gretz et~al.(2020)Gretz, Friedman, Cohen-Karlik, Toledo, Lahav,
  Aharonov, and
  Slonim}]{Gretz_Friedman_Cohen-Karlik_Toledo_Lahav_Aharonov_Slonim_2020}
Shai Gretz, Roni Friedman, Edo Cohen-Karlik, Assaf Toledo, Dan Lahav, Ranit
  Aharonov, and Noam Slonim. 2020.
\newblock \href {https://doi.org/10.1609/aaai.v34i05.6285} {A large-scale
  dataset for argument quality ranking: Construction and analysis}.
\newblock \emph{Proceedings of the AAAI Conference on Artificial Intelligence},
  34(05):7805--7813.

\bibitem[{He et~al.(2020)He, Liu, Gao, and Chen}]{he2020deberta}
Pengcheng He, Xiaodong Liu, Jianfeng Gao, and Weizhu Chen. 2020.
\newblock \href {http://arxiv.org/abs/2006.03654} {Deberta: Decoding-enhanced
  bert with disentangled attention}.

\bibitem[{Kapadnis et~al.(2021)Kapadnis, Patnaik, Panigrahi, Madhavan, and
  Nandy}]{KPA2021-31}
Manav~Nitin Kapadnis, Sohan Patnaik, Siba~Smarak Panigrahi, Varun Madhavan, and
  Abhilash Nandy. 2021.
\newblock Team enigma at argmining-emnlp 2021: Leveraging pre-trained language
  models for key point matching.
\newblock In \emph{Proceedings of the 8th Workshop on Argument Mining}, Online.
  Association for Computational Linguistics.

\bibitem[{Lan et~al.(2019)Lan, Chen, Goodman, Gimpel, Sharma, and
  Soricut}]{DBLP:journals/corr/abs-1909-11942}
Zhenzhong Lan, Mingda Chen, Sebastian Goodman, Kevin Gimpel, Piyush Sharma, and
  Radu Soricut. 2019.
\newblock \href {http://arxiv.org/abs/1909.11942} {{ALBERT:} {A} lite {BERT}
  for self-supervised learning of language representations}.
\newblock \emph{CoRR}, abs/1909.11942.

\bibitem[{Liu et~al.(2018)Liu, Saleh, Pot, Goodrich, Sepassi, Kaiser, and
  Shazeer}]{j.2018generating}
Peter~J. Liu, Mohammad Saleh, Etienne Pot, Ben Goodrich, Ryan Sepassi, Lukasz
  Kaiser, and Noam Shazeer. 2018.
\newblock \href {https://openreview.net/forum?id=Hyg0vbWC-} {Generating
  wikipedia by summarizing long sequences}.
\newblock In \emph{International Conference on Learning Representations}.

\bibitem[{Liu and Lapata(2019)}]{liu-lapata-2019-roberta}
Yang Liu and Mirella Lapata. 2019.
\newblock \href {https://doi.org/10.18653/v1/D19-1387} {Text summarization with
  pretrained encoders}.
\newblock In \emph{Proceedings of the 2019 Conference on Empirical Methods in
  Natural Language Processing and the 9th International Joint Conference on
  Natural Language Processing (EMNLP-IJCNLP)}, pages 3730--3740, Hong Kong,
  China. Association for Computational Linguistics.

\bibitem[{Liu et~al.(2019)Liu, Ott, Goyal, Du, Joshi, Chen, Levy, Lewis,
  Zettlemoyer, and Stoyanov}]{Liu2019RoBERTaAR}
Yinhan Liu, Myle Ott, Naman Goyal, Jingfei Du, Mandar Joshi, Danqi Chen, Omer
  Levy, M.~Lewis, Luke Zettlemoyer, and Veselin Stoyanov. 2019.
\newblock Roberta: A robustly optimized bert pretraining approach.
\newblock \emph{ArXiv}, abs/1907.11692.

\bibitem[{Lu et~al.(2020)Lu, Dong, and Charlin}]{lu-etal-2020-multi-xscience}
Yao Lu, Yue Dong, and Laurent Charlin. 2020.
\newblock \href {https://doi.org/10.18653/v1/2020.emnlp-main.648}
  {Multi-{XS}cience: A large-scale dataset for extreme multi-document
  summarization of scientific articles}.
\newblock In \emph{Proceedings of the 2020 Conference on Empirical Methods in
  Natural Language Processing (EMNLP)}, pages 8068--8074, Online. Association
  for Computational Linguistics.

\bibitem[{Phan et~al.(2021)Phan, Nguyen, Nguyen, and Doan}]{KPA2021-17}
Hoang~Viet Phan, Long~Tien Nguyen, Long~Duc Nguyen, and Khanh Doan. 2021.
\newblock Matching the statements: A simple and accurate model for key point
  analysis.
\newblock In \emph{Proceedings of the 8th Workshop on Argument Mining}, Online.
  Association for Computational Linguistics.

\bibitem[{Reimer et~al.(2021)Reimer, Luu, Henze, and Ajjour}]{KPA2021-6}
Jan~Heinrich Reimer, Thi Kim~Hanh Luu, Max Henze, and Yamen Ajjour. 2021.
\newblock Modern talking in key point analysis: Key point matching using
  pretrained encoders.
\newblock In \emph{Proceedings of the 8th Workshop on Argument Mining}, Online.
  Association for Computational Linguistics.

\bibitem[{Shapira et~al.(2021)Shapira, Pasunuru, Ronen, Bansal, Amsterdamer,
  and Dagan}]{shapira-etal-2021-extending}
Ori Shapira, Ramakanth Pasunuru, Hadar Ronen, Mohit Bansal, Yael Amsterdamer,
  and Ido Dagan. 2021.
\newblock \href {https://doi.org/10.18653/v1/2021.naacl-main.54} {Extending
  multi-document summarization evaluation to the interactive setting}.
\newblock In \emph{Proceedings of the 2021 Conference of the North American
  Chapter of the Association for Computational Linguistics: Human Language
  Technologies}, pages 657--677, Online. Association for Computational
  Linguistics.

\bibitem[{Syed et~al.(2020)Syed, El~Baff, Kiesel, Al~Khatib, Stein, and
  Potthast}]{syed-etal-2020-news}
Shahbaz Syed, Roxanne El~Baff, Johannes Kiesel, Khalid Al~Khatib, Benno Stein,
  and Martin Potthast. 2020.
\newblock \href {https://doi.org/10.18653/v1/2020.coling-main.470} {News
  editorials: Towards summarizing long argumentative texts}.
\newblock In \emph{Proceedings of the 28th International Conference on
  Computational Linguistics}, pages 5384--5396, Barcelona, Spain (Online).
  International Committee on Computational Linguistics.

\bibitem[{Toledo et~al.(2019)Toledo, Gretz, Cohen-Karlik, Friedman, Venezian,
  Lahav, Jacovi, Aharonov, and Slonim}]{toledo-etal-2019-automatic}
Assaf Toledo, Shai Gretz, Edo Cohen-Karlik, Roni Friedman, Elad Venezian, Dan
  Lahav, Michal Jacovi, Ranit Aharonov, and Noam Slonim. 2019.
\newblock \href {https://doi.org/10.18653/v1/D19-1564} {Automatic argument
  quality assessment - new datasets and methods}.
\newblock In \emph{Proceedings of the 2019 Conference on Empirical Methods in
  Natural Language Processing and the 9th International Joint Conference on
  Natural Language Processing (EMNLP-IJCNLP)}, pages 5625--5635, Hong Kong,
  China. Association for Computational Linguistics.

\bibitem[{Turpin and Scholer(2006)}]{mAP_reference}
Andrew Turpin and Falk Scholer. 2006.
\newblock \href {https://doi.org/10.1145/1148170.1148176} {User performance
  versus precision measures for simple search tasks}.
\newblock In \emph{Proceedings of the 29th Annual International ACM SIGIR
  Conference on Research and Development in Information Retrieval}, SIGIR '06,
  page 11–18, New York, NY, USA. Association for Computing Machinery.

\bibitem[{Wang and Ling(2016)}]{wang-ling-2016-neural}
Lu~Wang and Wang Ling. 2016.
\newblock \href {https://doi.org/10.18653/v1/N16-1007} {Neural network-based
  abstract generation for opinions and arguments}.
\newblock In \emph{Proceedings of the 2016 Conference of the North {A}merican
  Chapter of the Association for Computational Linguistics: Human Language
  Technologies}, pages 47--57, San Diego, California. Association for
  Computational Linguistics.

\bibitem[{Xu and Lapata(2020)}]{xu-lapata-2020-coarse}
Yumo Xu and Mirella Lapata. 2020.
\newblock \href {https://doi.org/10.18653/v1/2020.emnlp-main.296}
  {Coarse-to-fine query focused multi-document summarization}.
\newblock In \emph{Proceedings of the 2020 Conference on Empirical Methods in
  Natural Language Processing (EMNLP)}, pages 3632--3645, Online. Association
  for Computational Linguistics.

\bibitem[{Yang et~al.(2019)Yang, Dai, Yang, Carbonell, Salakhutdinov, and
  Le}]{NEURIPS2019_XLNet}
Zhilin Yang, Zihang Dai, Yiming Yang, Jaime Carbonell, Russ~R Salakhutdinov,
  and Quoc~V Le. 2019.
\newblock \href
  {https://proceedings.neurips.cc/paper/2019/file/dc6a7e655d7e5840e66733e9ee67cc69-Paper.pdf}
  {Xlnet: Generalized autoregressive pretraining for language understanding}.
\newblock In \emph{Advances in Neural Information Processing Systems},
  volume~32. Curran Associates, Inc.

\bibitem[{Zhang et~al.(2020)Zhang, Zhao, Saleh, and Liu}]{pmlr-v119-zhang20ae}
Jingqing Zhang, Yao Zhao, Mohammad Saleh, and Peter Liu. 2020.
\newblock \href {https://proceedings.mlr.press/v119/zhang20ae.html} {{PEGASUS}:
  Pre-training with extracted gap-sentences for abstractive summarization}.
\newblock In \emph{Proceedings of the 37th International Conference on Machine
  Learning}, volume 119 of \emph{Proceedings of Machine Learning Research},
  pages 11328--11339. PMLR.

\end{thebibliography}
\bibliographystyle{acl_natbib}

\appendix

\section{Appendix}
\begin{table}[ht!]
\begin{center}
\small

\begin{tabular}{c| c| c | c} 
 \hline

Model & Stance & Redundancy & Relevance\\
\hline
\barh & 95.46 & 42.73 & 28.63 \\ \hline
\mspl & 2.91 & 42.92 & 28.62 \\ \hline
\sohanpat & 1.37 & 4.47 & 0.0 \\ \hline
\peratham & 0.27 & 9.89 & 42.76 \\ \hline

\end{tabular}
\small
\end{center}
\caption{Bradley Terry scores for the comparative evaluation of the top models in \genTrack.
\label{app:bt_table}}
\end{table}

\begin{table*}[ht]
\centering
\small
\begin{tabular}{c|c|c|c|c|c|c|c}
\hline

\multirow{3}{*}{Model}&\multicolumn{2}{p{3cm} |}{Routine child vaccinations should be mandatory}&\multicolumn{2}{p{3cm} |}{Social media platforms should be regulated by the government}&\multicolumn{2}{p{3cm} |}{The USA is a good country to live in} & \multirow{3}{*}{Average} \\ 
\cline{2-7}
& pos& neg& pos& neg& pos& neg &\\
\hline
\mspl & 0.29 & 0.13 & 0.247 & 0.245 & 0.157 & 0.252 & 0.22\\
\hline
\vund & 0.328 & 0.144 & \textbf{0.0455} & 0.205 & \textbf{0.0585} & 0.153 & 0.156\\
\hline
\heinrich & 0.238 & 0.131 & 0.242 & 0.216 & 0.125 & 0.263 & 0.202\\
\hline
\sohanpat & 0.223 & 0.131 & 0.232 & \textbf{0.158} & 0.146 & \textbf{0.118} & 0.168\\
\hline
\viethoang & \textbf{0.181} & \textbf{0.105} & 0.0797 & 0.245 & 0.107 & 0.196 & 0.152\\
\hline
\barh & 0.431 & 0.172 & 0.236 & 0.205 & 0.121 & 0.261 & 0.238\\
\hline

\end{tabular}
\caption{Jansen-Shannon divergence for top five models and the state of the art model. The divergence is reported for each motion, and is also averaged. Best score for each model is bolded.
\label{tab:JS_divergence}}
\end{table*}

\begin{table*}[ht]
\centering
\small
\begin{tabular}{p{2cm}|p{10.3cm}|p{0.5cm}}

\hline
 model & \centering{key point} &  \% \\ \hline

\multirow{5}{*}{Expert} &Routine child vaccinations should be mandatory to prevent virus/disease spreading & 37\\ \cline{2-3}
&Children should not suffer from preventable diseases & 29\\ \cline{2-3}
&Child vaccination saves lives & 20\\ \cline{2-3}
&Routine child vaccinations are necessary to protect others & 8\\ \cline{2-3}
&Routine child vaccinations are effective & 6\\ \cline{2-3}
\hline \hline

\multirow{7}{*}{\barh} &to keep schools safe children must be vaccinated & 35\\ \cline{2-3}
&Vaccines help children grow up healthy and avoid dangerous diseases & 35\\ \cline{2-3}
&The use of child vaccines saves lives & 11\\ \cline{2-3}
&our children have the right to be vaccinated & 6\\ \cline{2-3}
&in this way, pandemic outbreaks are avoided & 5\\ \cline{2-3}
&After getting vaccinated, our immune system produces antibodies. & 4\\ \cline{2-3}
&Child vaccination should be mandatory to provide decent healthcare equally & 2\\  \hline

 \multirow{3}{*}{\mspl}&Vaccines help children grow up healthy and avoid dangerous diseases & 77\\ \cline{2-3}
&child vaccinations should be mandatory to provide decent health care to all. & 12\\ \cline{2-3}
&child vaccinations should be mandatory so our children will be safe and protected. & 9\\ \cline{2-3}
 \hline

 \multirow{4}{*}{\sohanpat}&Child vaccinations are necessary to protect against diseases in the child & 50\\ \cline{2-3}
&A mandatory vaccination is important to protect against diseases in the child & 25\\ \cline{2-3}
&Children should be vaccinated against diseases & 23\\ \cline{2-3}
&Children should be vaccinated against diseases as soon as possible & 2\\
 \hline

\multirow{16}{*}{\peratham} &It is important to protect the health of the next generation so to do that means the routine child vaccinations have to be mandatory. & 57\\ \cline{2-3}
&vaccines must be compulsory for children because in this way we prevent later diseases in infants & 11\\ \cline{2-3}
&Making routine child vaccinations mandatory would prevent the unnecessary deaths and suffering that currently result from the diseases that the vaccines prevent. & 8\\ \cline{2-3}
&child vaccinations should be mandatory so our children will be safe and protected. & 8\\ \cline{2-3}
&Child vaccination has to be mandatory since it takes care of our children to be immune to any type of disease so I think it has to be mandatory & 4\\ \cline{2-3}
&childhood vaccines should be mandatory because a child at an early age needs to strengthen their immune system and the state should be in the power to provide such protection & 4\\ \cline{2-3}
&If routine infallible vaccinations are mandatory, that way we avoid the spread and make our children immune from any disease. & 2\\ \cline{2-3}
&The main reason is that many diseases have been prevented thanks to vaccines, thus millions of lives have been saved. & 2\\ 
 \hline

\multirow{5}{*}{\pavincent}&vaccines have be able to protect diseases & 46\\ \cline{2-3}
&child are be used to prevent diseases & 33\\ \cline{2-3}
&parents vaccines should be safe for children & 7\\ \cline{2-3}
&vaccines harms risk of the child & 6\\ \cline{2-3}
&vaccines is important for protection & 6\\ 
 \hline

\end{tabular}

\caption{Expert and generated key points for \routineVaccines pro stance. Fraction is out of all arguments with a matched \kp{} (for expert we use the strict view).
\label{tab:key_points_full}}
\end{table*}

\end{document}